\documentclass[lettersize,journal]{IEEEtran}
\usepackage{amsmath,amsfonts}
\usepackage{algorithmic}
\usepackage{algorithm}
\usepackage{array}
\usepackage[caption=false,font=normalsize,labelfont=sf,textfont=sf]{subfig}
\usepackage{textcomp}
\usepackage{stfloats}
\usepackage{url}
\usepackage{verbatim}
\usepackage{graphicx}
\usepackage{cite}
\usepackage{subfiles}
\usepackage{amsfonts}

\usepackage{booktabs}
\usepackage{subfiles}
\usepackage{bbding}
\usepackage{url}
\usepackage{booktabs}
\usepackage{amssymb}
\usepackage{bbding}
\usepackage{pifont}
\usepackage{wasysym}
\usepackage{utfsym}
\usepackage{fontawesome}
\usepackage{bibentry}
\usepackage{array}
\usepackage{makecell}
\usepackage{color, colortbl}
\usepackage{multicol}
\usepackage{multirow}
\usepackage{mathrsfs}
\usepackage{bm}
\usepackage{hyperref}

\definecolor{lime}{HTML}{A6CE39}
\DeclareRobustCommand{\orcidicon}{
	\begin{tikzpicture}
	\draw[lime, fill=lime] (0,0) 
	circle [radius=0.16] 
	node[white] {{\fontfamily{qag}\selectfont \tiny ID}};
	\draw[white, fill=white] (-0.0625,0.095) 
	circle [radius=0.007];
	\end{tikzpicture}
	\hspace{-2mm}
}
\newcommand{\orcid}[1]{\href{https://orcid.org/#1}{\orcidicon}}

\begin{document}

\title{Gaze-guided Hand-Object Interaction Synthesis: Dataset and Method}

\author{

Jie Tian\orcid{0009-0005-8750-2031}, 
Ran Ji\orcid{0009-0001-2874-6475},
Lingxiao Yang\orcid{0009-0000-9917-4470}, 
Suting Ni\orcid{0009-0001-3779-1657}, 
Yuexin Ma\orcid{0000-0001-7237-988X}~\IEEEmembership{Member, IEEE}, 
Lan Xu\orcid{0000-0002-8807-7787}~\IEEEmembership{Member, IEEE}, 
Jingyi Yu\orcid{0000-0002-8580-0036}~\IEEEmembership{Fellow, IEEE}, 
Ye Shi\orcid{0000-0003-2302-4980}~\IEEEmembership{Member, IEEE}, 
Jingya Wang\orcid{0000-0003-1122-0634}~\IEEEmembership{Member, IEEE}

\thanks{Received 07 Jan 2025; revised 14 Dec 2025; accepted 28 Jan 2026. This work was supported by National Natural Science Foundation of China (62406195, 62303319), ShanghaiTech AI4S Initiative SHTAI4S202404, HPC Platform of ShanghaiTech University, and MoE Key Laboratory of Intelligent Perception and Human-Machine Collaboration (ShanghaiTech University), and Shanghai Engineering Research Center of Intelligent Vision and Imaging. This work was also supported in part by computational resources provided by Fcloud CO., LTD.}
\thanks{The authors are with the School of Information Science and Technology, ShanghaiTech University, Shanghai 201210, China (e-mail: \{tianjie2022, jiran2022, yanglx2023, nist2024, mayuexin, xulan1, yujingyi, shiye, wangjingya\}@shanghaitech.edu.cn).Corresponding author: Jingya Wang.}}

\makeatletter

\let\@oldmaketitle\@maketitle
\renewcommand{\@maketitle}{
   \@oldmaketitle
   \setcounter{figure}{0}
 \begin{center}
      \includegraphics[width=1.0\linewidth]{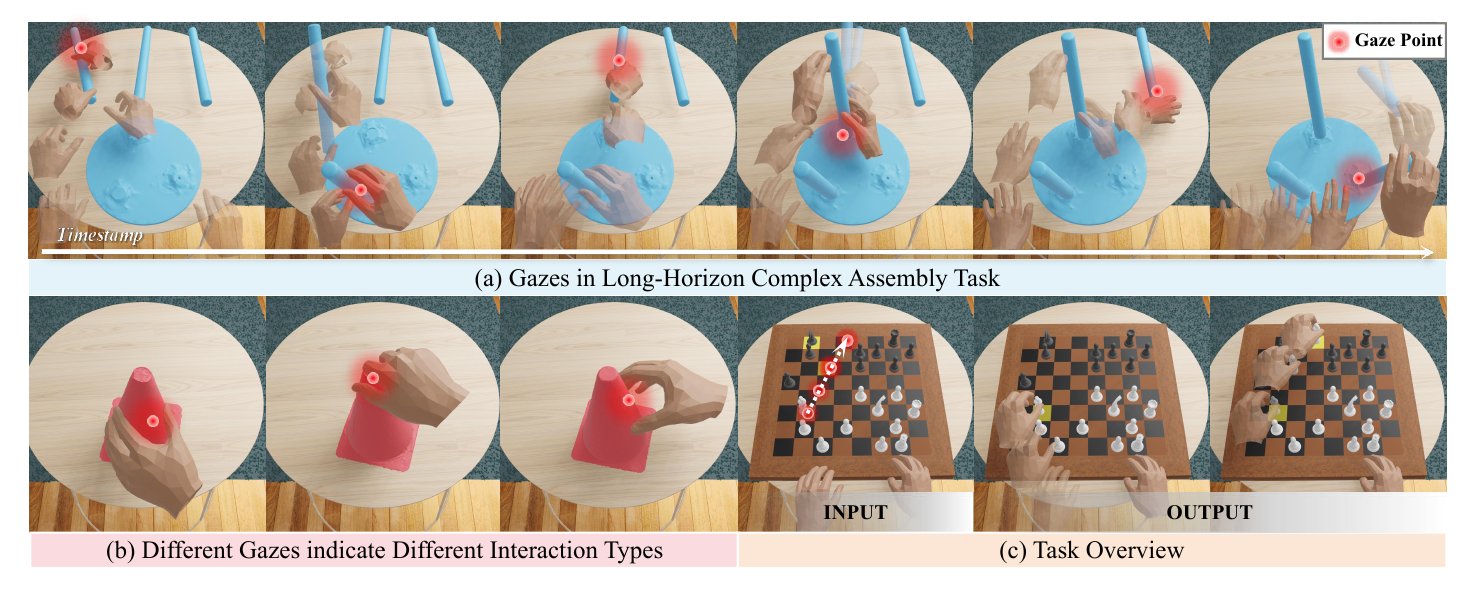}
 \end{center}
 \vspace{-2ex}
  \refstepcounter{figure}\normalfont Figure~\thefigure. 
Gaze plays a crucial role in hand-object interaction tasks. The subfigure (a) illustrates how gaze helps coordinate the hands, objects, and brain during complex table assembly. The subfigure (b) shows how gaze affects attention allocation and grasp strategies. Building on these observations, we introduce the gaze-guided hand-object interaction synthesis task, as depicted in the subfigure (c), which uses gaze data as input to generate hand-object interactions that align with human intentions. 
\vspace{-1cm}
  \label{fig:teaser}}
\makeatother

\maketitle

\begin{abstract}
Gaze plays a crucial role in revealing human attention and intention, particularly in hand-object interaction scenarios, where it guides and synchronizes complex tasks that require precise coordination between the brain, hand, and object. Motivated by this, we introduce a novel task: Gaze-Guided Hand-Object Interaction Synthesis, with potential applications in augmented reality, virtual reality, and assistive technologies. To support this task, we present GazeHOI, the first dataset to capture simultaneous 3D modeling of gaze, hand, and object interactions. This task poses significant challenges due to the inherent sparsity and noise in gaze data, as well as the need for high consistency and physical plausibility in generating hand and object motions. To tackle these issues, we propose a stacked gaze-guided hand-object interaction diffusion model, named GHO-Diffusion. The stacked design effectively reduces the complexity of motion generation. We also introduce HOI-Manifold Guidance during the sampling stage of GHO-Diffusion, enabling fine-grained control over generated motions while maintaining the data manifold. Additionally, we propose a spatial-temporal gaze feature encoding for the diffusion condition and select diffusion results based on consistency scores between gaze-contact maps and gaze-interaction trajectories. Extensive experiments highlight the effectiveness of our method and the unique contributions of our dataset. More details in https://takiee.github.io/gaze-hoi/.
\end{abstract}


\section{Introduction}

\IEEEPARstart{G}{aze} serves as a significant behavioral signal, directly reflecting human attention distribution and cognitive processes \cite{con1,con3,con4grasp}.
Extensive research has been conducted on gaze, including gaze estimation \cite{ge1,ge2,ge3,ge4,Lv2021ImprovingDG,Zhang2024DomainConsistentAU,Lou2024PredictingRG} and human action prediction based on gaze \cite{mogaze,gimo,actionsense,meccano,gazedrive}.
While these studies have explored gaze in various applications, they have primarily focused on broad human motions, with limited attention given to hand movements and detailed dynamic interactions within fine-grained environments.
 Furthermore, gaze is particularly crucial in hand-object interactions, as it reveals "\textit{when, where, and how}" the user interacts with the object. Figure \ref{fig:teaser}(a) illustrates how gaze supports long-horizon synthesis for complex tasks, while Figure \ref{fig:teaser}(b) shows that gaze indicates how humans interact with objects using different parts and interaction types. By effectively identifying objects and areas of focus, gaze enables the seamless integration of human intentions and motions, ensuring precise coordination and consistency between the brain, hands, and objects. This coordination is essential for executing complex interactions, such as grasping, manipulating, or positioning objects, thereby enhancing both the efficiency and accuracy of these motions.

Motivated by this, we propose a novel task: Gaze-guided Hand-Object Interaction Synthesis, as shown in Figure \ref{fig:teaser}(c). This task leverages gaze to naturally generate motions that align with human intent, eliminating the need for additional textual annotations or motion references. This approach is particularly well-suited for integration into VR/AR environments and assistive technologies. For example, in VR/AR settings, gaze-guided technology allows users to interact with virtual objects more intuitively by accurately predicting their intentions without the need for physical controllers. Additionally, it benefits individuals with disabilities by allowing them to control interactions through eye movements, enhancing their ability to express intentions and perform tasks

Existing hand-object datasets\cite{ho3d,dexycb,h2o,oakink,hoi4d,arctic,taco,neuraldome,zhao2023m,zhan2024oakink2} are limited to interactions between hands and objects, often neglecting gaze information. To fill this gap, we introduce GazeHOI, the first dataset that simultaneously integrates 3D modeling of gaze, hand, and object interactions. GazeHOI presents significant challenges due to the diverse shapes and sizes of objects, ranging from small items like chess pieces to larger, assemblable furniture. Additionally, the dataset covers a wide range of complex tasks, including repositioning objects, selecting specific targets within cluttered environments, organizing disordered items, and assembling furniture.

However, leveraging gaze data for generating hand and object motions presents significant challenges. The inherent sparsity and noise in gaze data make it difficult to reliably capture and interpret user intentions. Additionally, achieving high consistency and physical plausibility in the generated motions is a complex task, as it requires accurate synchronization between gaze input and corresponding hand-object interactions. To address these challenges, we propose a novel approach: a stacked gaze-guided hand-object interaction diffusion model, named GHO-Diffusion. We decouple the task into two stages: gaze-activated object dynamics synthesis and object-driven hand kinematic synthesis. Moreover, we introduce HOI-Manifold Guidance during the sampling stage of GHO-Diffusion. This technique allows for fine-grained control over the generated motions, ensuring they remain within the data manifold and adhere to the natural constraints of hand-object interactions.
Additionally, to further leverage gaze information, we propose a spatial-temporal gaze feature encoding approach for the diffusion condition. This method captures both spatial and temporal aspects of gaze data, enhancing the diffusion process with detailed representations of gaze interactions. We also select diffusion results based on consistency scores between gaze-contact maps and gaze-interaction trajectories.
Our contributions can be summarized as follows:

\noindent\textbf{1) A Novel Task and Dataset.} We introduce the novel task of gaze-guided hand-object interaction synthesis, which utilizes gaze to generate natural interaction motions that align with human intent. To support this task, we present GazeHOI, the first dataset incorporating gaze information for hand-object interactions. 
    
\noindent\textbf{2) GHO-Diffusion Model.} We develop the Gaze-guided Hand-Object interaction Diffusion model, GHO-Diffusion, which employs stacked diffusion models to reduce the complexity of motion modeling. Furthermore, we integrate the HOI-Manifold Guidance during the inference phase of GHO-Diffusion, providing fine-grained control over generated motions while maintaining the data distribution learned by the diffusion model. 
    
\noindent\textbf{3) Gaze-Guided Feature Encoding and Motion Selection.} We propose a gaze-guided spatial-temporal feature encoding as the diffusion condition, coupled with a mechanism to select diffusion results based on consistency scores between gaze-contact maps and gaze-interaction trajectories. This approach ensures that the generated motions are precisely aligned with the given gaze input. 
\begin{figure*}[!ht]
  \centering
   \includegraphics[width=1\linewidth]{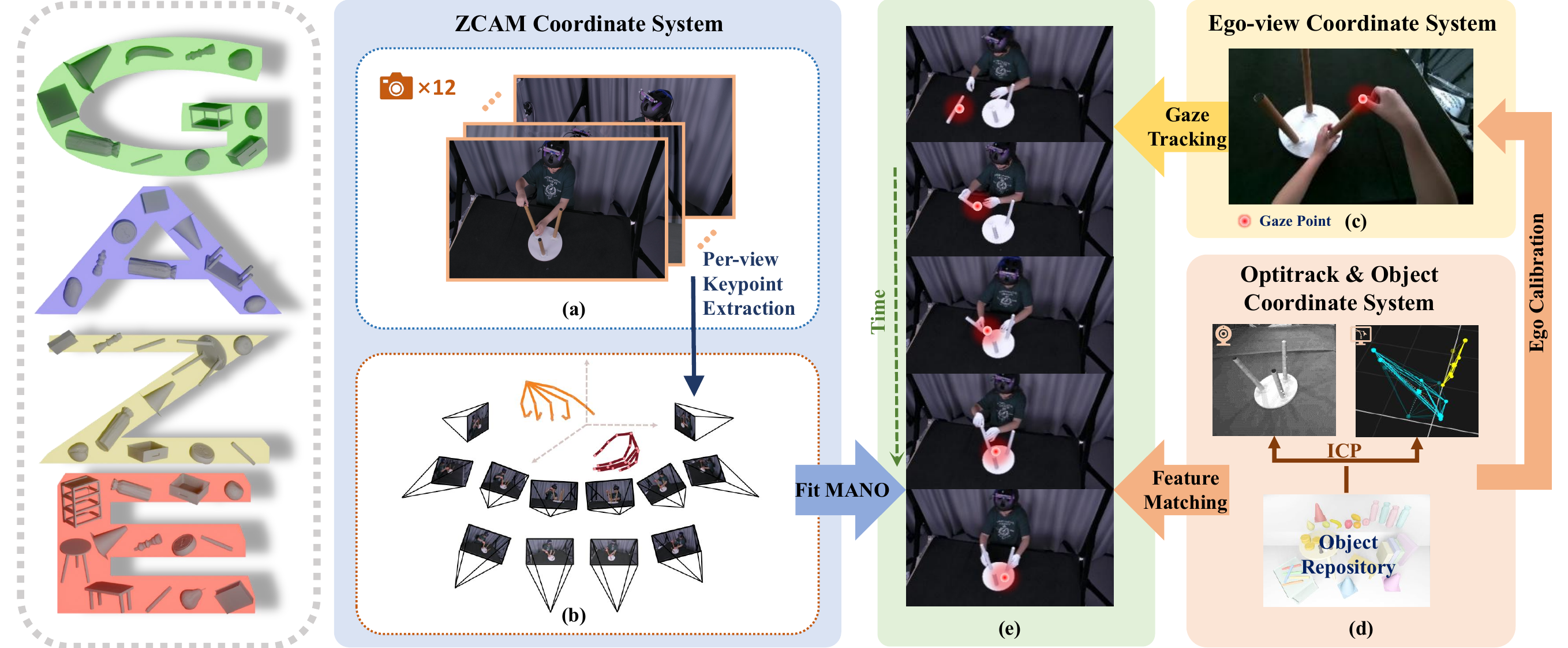}
   \caption{Automatic data processing pipeline: (a) raw 12-view images; (b) 3D hand joint estimation via MediaPipe \cite{mediapipe} and triangulation; (c) ego-view gaze; (d) object scanning and marker tracking; (e) final hand-object motion.}
   \vspace{-2ex}
   \label{fig:dataset}
\end{figure*}

\section{Related Work}
\subsection{Gaze for Perception}
Gaze is a key behavioral signal for inferring intentionality, attention, and cognitive processes. Early research, limited by equipment accuracy, focused on third-person gaze estimation \cite{ge1, Lv2021ImprovingDG,Zhang2024DomainConsistentAU,Lou2024PredictingRG,ge2,ge3,ge4} to predict human attention regions. Recent tracking advancements \cite{Chen20233DFR} have facilitated diverse first-person datasets \cite{mogaze,gimo,actionsense,meccano,gazedrive,holoassist,aria,egobody} that leverage gaze for contextual tasks like ego-trajectory inference \cite{gazedrive} or action labeling \cite{liu2020forecasting}. However, existing gaze-motion benchmarks \cite{mogaze,gimo,actionsense,holoassist,aria,egobody} largely overlook the link between gaze and hand-object interaction (HOI)—essential for fine-grained intent inference and realism in VR.
Specifically, current datasets omit either hand \cite{mogaze,aria} or object poses \cite{gimo,holoassist,actionsense,egobody}, as shown in Table \ref{tab:gaze data}. Methodologically, most studies focus on human motion prediction \cite{gazemodiff,gazemotion,gimo}, treating gaze as auxiliary input to fusion encoders and relying heavily on historical sequences. These approaches fail to model HOI precision, such as realistic contact, penetration avoidance, and plausible physical dynamics. To bridge these gaps, we present the first dataset capturing 3D gaze, hand, and object interactions simultaneously. We introduce a novel task: gaze-guided HOI synthesis from an initial state without relying on historical motions. Our method utilizes gaze-guided spatial-temporal feature encoding as a diffusion condition and selects results based on gaze-contact and trajectory consistency scores to effectively capture the intentionality inherent in gaze.

\begin{table}[!t]
    \centering
    \caption{comparison with Existing gaze datasets}
    \vspace{-2ex}
    \begin{tabular}{cccc}
    \toprule[1.5pt]
         &Gaze     &Object pose  &Hand pose\\
     \midrule
        MoGaze\cite{mogaze} & $\checkmark$ 
 & $\checkmark$ & $\times$ \\
         GIMO\cite{gimo} & $\checkmark$ & $\times$ & $\checkmark$   \\
        ActionSense\cite{actionsense} & $\checkmark$ 
 & $\times$ & $\checkmark$  \\
 MECCANO\cite{meccano} & $\checkmark$ & $\times$ & $\times$  \\
         HoloAssist\cite{holoassist} & $\checkmark$ & $\times$ & $\checkmark$ \\
         ADT\cite{aria} & $\checkmark$ & $\checkmark$ & $\times$ \\
         EgoBody\cite{egobody} & $\checkmark$ & $\times$ & $\checkmark$  \\
    \midrule
         \textbf{GazeHOI}& $\checkmark$ & $\checkmark$ & $\checkmark$   \\
    \bottomrule[1.5pt] 
    \end{tabular}
    \label{tab:gaze data}
    \vspace{-4ex}
\end{table}
\subsection{Hand-Object Interaction Dataset}
Numerous datasets \cite{ho3d,dexycb,h2o,oakink,hoi4d,arctic,taco,neuraldome,zhao2023m,zhan2024oakink2} dedicated to hand-object interactions have been developed to facilitate the capture of hand and object poses and to deepen our understanding of these interactions, as shown in Table \ref{tab:hoi_dataset}. 
Most datasets only capture simple interaction patterns,like grasping and moving, which is common but not comprehensive enough. HOI4D \cite{hoi4d} and ARCTIC \cite{arctic} explore the interaction of articulated objects.
TACO \cite{taco} and OakInk2 \cite{zhan2024oakink2} captures scenes of simultaneous interactions with multiple objects. 
Furthermore, existing datasets focus exclusively on the relationship between hands and objects, neglecting the role of gaze. To address this gap, we introduce a comprehensive egocentric dataset enriched with detailed gaze annotations. This dataset captures the dynamics of hand-object interactions across a wide range of tasks, from arranging chess pieces to assembling furniture — scenarios not covered in existing datasets. Our goal is to bridge the shortcomings of current datasets by providing a resource that reflects the complexity of real-world interactions and emphasizes the importance of gaze in predicting human intent and managing multi-object manipulation.

\begin{table}[!t]
    \centering
    \caption{Comparison with existing 4D hand-object interaction datasets.}
    \scriptsize
    {
        \begin{tabular}{ccccccccc}
        \toprule[1.5pt]
             &\shortstack{Gaze}&\shortstack{Ego}&\shortstack{Multi-obj}&\shortstack{Complex} &\shortstack{\#views}&\shortstack{\#images}\\
         \midrule
            HO-3D \cite{ho3d} & $\times$ & $\times$& $\times$ & $\times$  &1-4 &78K \\
            DexYCB \cite{dexycb}& $\times$& $\times$ & $\times$ & $\times$  &8 &582K\\
            H2O \cite{h2o}& $\times$ &$\checkmark$ & $\times$& $\times$   & 5 &571K\\
             OakInk \cite{oakink}& $\times$ & $\times$& $\times$ & $\times$  &4  &230K\\
             HOI4D \cite{hoi4d}& $\times$&$\checkmark$ & $\times$& $\times$   &2 &2.4M\\
             ARCTIC \cite{arctic}& $\times$&$\checkmark$ & $\times$  & $\times$  &9&2.1M \\
             TACO \cite{taco}& $\times$  & $\checkmark$ & $\checkmark$& $\checkmark$ &13 &5.2M\\
             OAKINK2 \cite{zhan2024oakink2} & $\times$ & $\checkmark$ & $\checkmark$ & $\checkmark$ &4 &4.01M\\
        \midrule
             \textbf{GazeHOI}& $\checkmark$& $\checkmark$& $\checkmark$& $\checkmark$ &13&3.2M\\
         \bottomrule[1.5pt]
        \end{tabular}
    }
    
    \label{tab:hoi_dataset}
    \vspace{-0.4cm} 
\end{table}
\vspace{-3ex}
\subsection{Hand-Object Motion Synthesis}
In the field of hand-object motion synthesis, prior efforts have mainly centered on frame-wise generation \cite{hdo_1,hdo_2,liu2023contactgen,contactopt}. For dynamic hand-object interaction, previous methods typically rely on motion references, such as trajectories \cite{hdo_3, ho_diff_1,grip,gears,omomo,hdo_4} pr grasp references \cite{hdo_6}, or only focus on approaching stage \cite{ho_vae_1, ho_vae_2}. 
Recently, there are several works that study the hand-object interaction generation directly based on text \cite{text2hoi, diffh2o},  significantly reducing the reliance on specific conditions.
Building on advancements in 3D gaze acquisition technology, our work further relaxes these constraints by generating hand-object interactions directly from gaze data. Compared to text, gaze-based generation aligns more closely with the user's intent, resulting in more intuitive hand-object interactions.

\section{GazeHOI Dataset}

Existing datasets primarily focus on the interaction between hands and objects, overlooking the crucial role of gaze. However, a comprehensive understanding of interaction requires coordinated hand-eye-object cooperation. To address this gap, we present GazeHOI, the first interaction dataset that simultaneously incorporates 3D modeling of gaze, hand, and object interactions. 
GazeHOI encompasses 237K frames with an average duration of 19.1 seconds and 2-4 objects per sequence, further divisible into 1378 subsequences. 
We collect 33 different objects with various shape and size, the smallest of which is a chess piece with a height of 7cm and a mere diameter of 2cm. Each sequence is accompanied by a task-level description,  which varies in complexity, and includes repositioning objects to another location, selecting specific targets from cluttered environments, organizing items from disarray into order, and assembling furniture. This rich variety aims to provide comprehensive insights into the intricate dynamics of interactive behaviors.

\vspace{-2ex}
\subsection{Dataset Hardware Setup}

To construct the GazeHOI dataset, we developed a data collection platform encompassing a 1.6m $\times$ 1.2m area, equipped with a multi-camera system. This platform comprises four distinct equipment systems. The ZCAM system includes 12 ZCAM E2 cameras that have been synchronized through hardware, capturing RGB signals at a resolution of 3840 x 2160 at 30fps, which are utilized for subsequent hand pose acquisition. The Optitrack system consists of 8 hardware-synchronized Optitrack px13W cameras, which use infrared to capture the positions of reflective markers at a frame rate of 120fps, facilitating the tracking of object poses. The Scanner system, comprising two EinScan Pro 2X and EinScan SE scanners, is dedicated to collecting geometric information about objects. The Ego system, primarily aimed at gathering gaze data, employs a Pupil Core connected to a RealSense D455, simultaneously recording gaze information along with RGBD data from an ego-centric perspective. We devise a sophisticated workflow to achieve coordinate system registration among the 4 equipment systems.
Each system is tasked with collecting specific types of data, and we have devised a sophisticated workflow to achieve coordinate system registration among the 4 equipment systems through reflective markers. We attached eight 10mm diameter markers on the surface of the RealSense and scanned it to obtain its 3D model. By annotating the position of each marker in every RGB viewpoint and triangulating these positions, we could ascertain the location of markers within each system, thereby aligning the coordinate systems of multiple devices.

\vspace{-2ex}
 \subsection{Data Collection}
We recruited 10 volunteers (5 males, 5 females) and provided them with comprehensive scripts detailing each task. Each task was performed over multiple iterations to capture natural gaze transitions. In unfamiliar settings, participants tend to scan their surroundings to understand the environment, leading to longer gaze durations on task-relevant objects. As familiarity increases, gaze becomes more focused and durations decrease. This protocol ensures the authenticity of our data by accurately reflecting real-world gaze behavior and adaptation during goal-oriented interactions.
\vspace{-2ex}
\subsection{Data Annotation}

\subsubsection{3D Hand Pose} 
12 synchronized ZCAM cameras provide an exhaustive 3D perspective of the hand (Fig. \ref{fig:dataset}a). We track 2D hand keypoints via MediaPipe \cite{mediapipe}, which are then triangulated into 3D global coordinates (Fig. \ref{fig:dataset}b). Finally, we optimize MANO \cite{mano} parameters by enforcing joint alignment and realistic anatomical constraints.

\subsubsection{Object 6D Pose} We focus on rigid objects, determining 6D poses by tracking at least eight 3mm reflective markers per object via Optitrack. Markers are strategically placed to minimize interaction interference and occlusion. After capturing precise geometries via a 3D scanner, we employ the Iterative Closest Point (ICP) algorithm to convert Optitrack reflection points into 6D poses (Fig. \ref{fig:dataset}d).
\subsubsection{Gaze Acquisition}
The eye tracker  Pupil Core provides 3D egocentric gaze data. To unify gaze, hand, and object data in a single coordinate system, we perform ego-calibration by attaching markers to the egocentric camera. Post-processing, including filtering short-duration gaze and smoothing, is applied to enhance data quality and reduce noise (Fig. \ref{fig:dataset}c).

\section{Method}

\begin{figure*}[!t]
\centering 
  \begin{minipage}{\textwidth}
    \centering
    \includegraphics[width=1\textwidth]{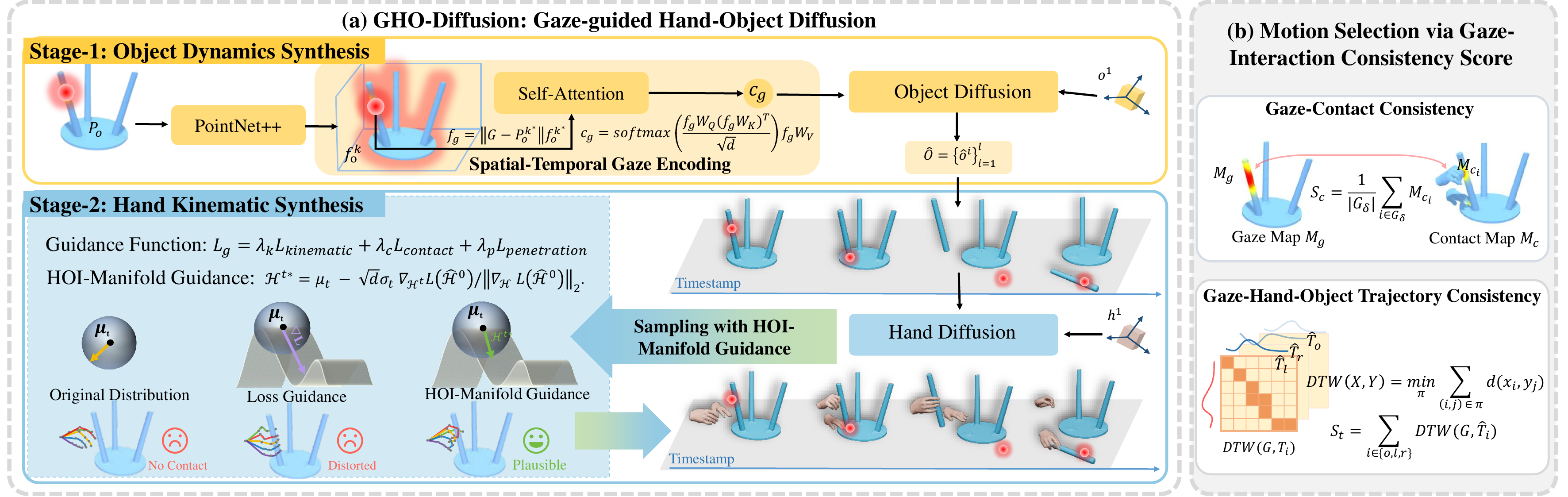}
  \end{minipage}
\caption{Pipeline overview. The subfigure (a) illustrates the stacked framework of the gaze-guided hand-object interaction diffusion model, GHO-Diffusion. The subfigure (b) shows the gaze-interaction consistency score for motion selection.
} 
\vspace{-2ex}
\label{method}
\end{figure*}

\subsection{Problem Definition}\label{task definition}
We define a gaze-consistent hand-object interaction sequence of length \( l \) as \( \mathbf{S} = \langle \mathbf{G}, \mathbf{H}, \mathbf{O} \rangle \), where
\( \mathbf{G} = \{\mathbf{g}_i\}_{i=1}^{l} \) represents the gaze sequence, with each \( \mathbf{g}_i \in \mathbb{R}^3 \) denoting the gaze point in the \( i \)-th frame.
 \( \mathbf{H} = \{\mathbf{h}_l^i, \mathbf{h}_r^i\}_{i=1}^{l} \) denotes the hand motion sequences, comprising left-hand motions \( \mathbf{h}_l^i \) and right-hand motions \( \mathbf{h}_r^i \). Each hand motion \( \mathbf{h}^i = \{ \mathbf{R}_h^i, \mathbf{T}_h^i, \boldsymbol{\theta}^i, \boldsymbol{\beta} \} \in \mathbb{R}^{61} \) follows the MANO \cite{mano} parameterization, where \( \mathbf{R}_h^i \) is the rotation, \( \mathbf{T}_h^i \) is the translation,  \( \boldsymbol{\theta}^i \) represents the pose and \( \boldsymbol{\beta} \) represents shape parameters; \( \mathbf{O} = \{\mathbf{o}^i\}_{i=1}^{l} \) is the object motion sequence, where each \( \mathbf{o}^i = \{\mathbf{R}_o^i, \mathbf{T}_o^i\} \in \mathbb{SE}(3) \) denotes the object's pose, with \( \mathbf{R}_o^i \) as the rotation and \( \mathbf{T}_o^i \) as the translation.
Additionally, we define the object geometry as a set of vertices \( \mathbf{P}_o = \{\mathbf{p}^n_o\}_{n=1}^{N} \in \mathbb{R}^{N \times 3} \), where \( N \) represents the number of object vertices.
Given the input \( \mathcal{I} = \{\mathbf{G}, \mathbf{h}^1_l, \mathbf{h}^1_r, \mathbf{o}^1, \mathbf{P}_o\} \), our objective is to derive a subsequent hand-object interaction sequence $\langle  \mathbf{H}, \mathbf{O} \rangle $ that is not only consistent with the gaze but also exhibits natural and realistic interactions between the hand and the object.

\subsection{GHO-Diffusion: Gaze-guided Hand-Object Diffusion}\label{4.3}

GHO-Diffusion decouples the complex hand-object interaction into two stacked diffusion stages, Object Dynamics Synthesis with Spatial-Temporal Gaze Encoding and Hand Kinematic Synthesis with HOI-Manifold Guidance as shown in Figure \ref{method}(a). 
This decoupling strategy reduces modeling complexity while effectively capturing both global spatial-temporal positioning and the intricate details of hand-object interaction motions. 

The diffusion model contains a fixed forward process $q(\mathbf{x}_t | \mathbf{x}_0) = \mathcal{N}(\sqrt{\alpha_t}\mathbf{x}_0,(1-\alpha_t)\mathbf{I})$ which gradually adds Gaussian noise to clean data $\mathbf{x}_0$ until it becomes pure noise $\boldsymbol{\epsilon} \sim \mathcal{N}(\mathbf{0},\mathbf{I})$ and train a denoiser $p_\theta(\mathbf{x}_t,t,\mathbf{c})$ to gradually denoise $\mathbf{x}_T \sim \mathcal{N}(\mathbf{0},\mathbf{I})$ to generate $\mathbf{x}_0$ in the reverse process. Instead of predicting $\boldsymbol{\epsilon}_t$ at each time step $t$, we follow MDM \cite{MDM} to predict $\hat{\mathbf{x}}_0$ and optimizing $p_\theta(\mathbf{x}_t,t,\mathbf{c})$ using the simple objective function:
\begin{equation}
    \label{eq:simple}
    \mathcal{L}_{\text{simple}} = \mathbb{E}_{{\mathbf{x}_0} \sim q(\mathbf{x}_0 | \mathbf{c}), t \sim [1, T]} ||\mathbf{x}_0 - p_\theta(\mathbf{x}_t, t, \mathbf{c})||_2^2.
\end{equation}

\subsubsection{Stage-1: Object Dynamics Synthesis with Spatial-Temporal Gaze Encoding.} 
The gaze point locations are closely related to the object surface vertices, and since object motions have fewer degrees of freedom compared to hand motions, so we generate the object motions first.
Gaze is a sparse and noisy signal, making it challenging to encode effectively.
Drawing inspiration from GIMO \cite{gimo}, we design a gaze-guided spatial-temporal feature encoding module for object diffusion. We define 
$\mathbf{f}^k_o$
as the $k$-th local per-point object spatial feature extracted by PointNet++ \cite{qi2017pointnet++}. Then according to the spatial relationship between gaze and object points, we can calculate gaze spatial feature: 
\begin{equation}
    \mathbf{f}_g = \lVert \mathbf{G} - \mathbf{P}^{k^*}_o \rVert^{-1} \mathbf{f}_o^{k^*},
\end{equation}
where $k^* = \arg \min_{k=1}^{N} \lVert \mathbf{G} - \mathbf{P}^{k}_o \rVert$, representing the index of the object points that is closest to the gaze point in $i$-th frame.
Subsequently, these spatial features are processed through a self-attention mechanism to extract the final spatial-temporal condition features: 
\begin{equation}
    \mathbf{c}_g = \text{softmax}\left(\frac{\mathbf{f}_g \mathbf{W}_Q (\mathbf{f}_g \mathbf{W}_K)^\top}{\sqrt{d}}\right) \mathbf{f}_g \mathbf{W}_V,
\end{equation}
where \( \mathbf{W}_Q \), \( \mathbf{W}_K \), and \( \mathbf{W}_V \) are trainable weight matrices, and \( d \) is the dimension of the feature.

 We leverage the gaze spatial-temporal features $\mathbf{c}_g$, along with the object geometry $\mathbf{P}$ and initial object pose $\mathbf{o}^1$  as stage-1 condition and employ the objective function  in Equation \ref{eq: stage1} as Stage-1 training loss to generate object motions $\hat{\mathbf{O}}$:
\begin{equation}
\label{eq: stage1}
     \mathcal{L}_\text{stage1}= \lambda_\alpha\mathcal{L}_\text{simple} + \lambda_v \mathcal{L}_\text{verts} + \lambda_t \mathcal{L}_\text{trans} + \lambda_s \mathcal{L}_\text{smooth}.
\end{equation}
where $\mathcal{L}_\text{simple}$ is defined by Equation \ref{eq:simple}.

The object translation loss $\mathcal{L}_\text{trans}$ is defined as:
\begin{equation}
    \mathcal{L}_\text{trans} = \sum\nolimits_{i=1}^L ||\mathbf{T}_o^i -\hat{\mathbf{T}}_o^i||_2,
\end{equation}
which enforces a constraint on the object’s position.
The object vertices loss $\mathcal{L}_\text{verts}$ is defined as:
\begin{equation} 
    \mathcal{L}_\text{verts} = \sum\nolimits_{i=1}^l ||(\mathbf{R}_o^i(\mathbf{P})+\mathbf{T}_o^i )- (\hat{\mathbf{R}}_o^i(\mathbf{P})+\hat{\mathbf{T}}_o^i)||_2,
\end{equation}
which enforces constraints on the object's surface geometry, promoting finer details and ensuring more precise alignment.
The smoothness loss $\mathcal{L}_\text{smooth}$ is defined as:
\begin{equation}
    \mathcal{L}_\text{smooth} = || \hat{\mathbf{O}}^{1:l} -  \hat{ \mathbf{O}}^{0:l-1}||_2,
\end{equation}
which promotes the continuity and fluidity of the generated motion. By minimizing the difference between consecutive motion frames, this loss ensures that the generated object motion is smooth and avoids abrupt transitions between frames.
\subsubsection{Stage-2: Hand Kinematic Synthesis with HOI Guidance} 
In this stage, we leverage the object motions $\hat{\mathbf{O}}$ generated in Stage-1 to synthesize the corresponding hand motions $\hat{\mathbf{H}}$ through a hand diffusion process. Hand motion synthesis presents greater complexity compared to object motion synthesis, primarily because hand movements are highly articulated and require more stringent constraints to ensure natural and physically plausible outcomes.
 To address this, we adopt a canonicalized hand-object interaction representation: ${\mathcal{H}} =[\mathbf{J}, \mathbf{C}, \mathbf{F}, \mathbf{v}, \mathbf{a}]$. Here, hand poses are represented by joint positions $\mathbf{J}$, as neural networks typically learn translations more effectively than rotations. The contact flag $\mathbf{C} \in \mathbb{R}^{42}$ encodes the contact status of each joint, while the offset vector $\mathbf{F}$ represents both hand-object and inter-hand spatial displacements. The velocity vector $\mathbf{v}$ captures both linear and angular velocities for each hand, including their relative velocities, and the acceleration vector $\mathbf{a}$ is similarly defined.
To generate the hand motions $\hat{\mathbf{H}}$, we condition the synthesis on the generated object motions $\hat{\mathbf{O}}$, alongside object geometry $\mathbf{P}$ and the initial hand pose $\mathbf{h}^1$ in Stage-2. For training, we employ the objective function from Equation \ref{eq:simple} and introduce a bone length loss $\mathcal{L}_\text{bone}$ to maintain the structural integrity of the hand:
\begin{equation}
\mathcal{L}_\text{stage2} = \lambda_\beta \mathcal{L}_\text{simple} + \lambda_b \mathcal{L}_\text{bone}.
\end{equation}
We define $\mathcal{B}(\cdot)$as a function that calculates the bone lengths, taking the hand joints $\mathbf{J}$ as input. The bone length loss $\mathcal{L}_\text{bone}$ is defined as: 
\begin{equation}
    \mathcal{L}_\text{bone} = ||\mathcal{B}(\mathbf{J}) - \mathcal{B}(\hat{\mathbf{J}})||_2.
\end{equation}
Since hand diffusion effectively captures the overall distribution of hand-object interactions, it falls short of achieving precision in finer details, such as maintaining contact or avoiding penetrations. Therefore, we introduce HOI-Manifold Guidance 
to strengthen the physics constraint while preserving the original data distribution. 
We define three guidance functions from the perspective of  Kinematic, Contact and Penetration.
\paragraph{\textbf{Hand-Object Kinematic Guidance.}} We define redundancy parameters to enhance the reliability of hand motions while using only the hand joints for subsequent steps. The kinematic consistency guidance is designed for test-time adaptation, ensuring that the generated parameters 
remain consistent with those recalculated from the generated hand joints and object motion conditions. The loss function is as follow, 
\begin{equation} 
\mathcal{L}_{\text{kinematic}} =   \| \hat{\mathbf{F}} - \tilde{\mathbf{F}} \|_2 + \|\hat{\mathbf{v}} - \tilde{\mathbf{v}} \|_2  + \| \hat{\mathbf{a}} - \tilde{\mathbf{a}}\|_2. 
\end{equation}
\paragraph{\textbf{Hand-Object Contact Guidance.}} 
We define a contact loss to minimize the distance between hand joints that are near the object's surface but not yet in contact: 
\begin{equation}
\mathcal{L}_{\text{contact}} =  \frac{\sum_{i=1}^{l}  d(\hat{\mathbf{J}}^i, \mathbf{P}_o^i) \cdot  \mathbb{I}(d(\hat{\mathbf{J}}^i, \mathbf{P}_o^i) < \tau) )}{\sum_{i=1}^{l} \mathbb{I}(d(\hat{\mathbf{J}}^i, \mathbf{P}_o^i) < \tau) + \epsilon},
\end{equation}
where \(d(\hat{\mathbf{J}}^i, \mathbf{P}_o^i)\) is the closest distance between hand joint \(\hat{\mathbf{J}}^i\) and object surface $\mathbf{P}_o^i$ in \(i\)-th frame, \(\tau\) is the contact threshold, \(\mathbb{I}\) is the indicator function, and \(\epsilon\) is a small constant to prevent division by zero.

\paragraph{\textbf{Hand-Object Penetration Guidance.}}
We calculate the dot product between the hand-object offset and the object surface normal to measure the penetration as follows,
\begin{equation}
\mathcal{L}_{\text{penetration}} = - \sum\nolimits_{i=1}^{l} \text{min}(0,  \tilde{\mathbf{F}}^i \cdot \mathbf{n}^i + \eta),
\end{equation}
where \(\tilde{\mathbf{F}}^i\) is the recalculated offset vector from the generated hand joint \(\hat{\mathbf{J}}^i\) to the object surface $\mathbf{P}_o^i$ in \(i\)-th frame, \(\mathbf{n}^i\) is the normal vector at the closest object point in \(i\)-th frame, and \(\eta\) is a small constant that defines the acceptable penetration threshold.
The overall guidance loss function is
\begin{equation}
    \mathcal{L}_g =  \lambda_k \mathcal{L}_{\text{kinematic}} + \lambda_c \mathcal{L}_{\text{contact}} + \lambda_p \mathcal{L}_{\text{penetration}}.
\end{equation}

The naive guidance strategy, which directly adds the computed gradient $\nabla_{{\mathcal{H}}^t} \mathcal{L}(\hat{\mathcal{H}}^0)$ to the sample, may misalign the final motions with the original data distribution, resulting in unnatural motions. 
Therefore, we aim to develop a guidance strategy that can achieve guidance targets while preserving the original distribution. 
Recently, a conditional diffusion model with the Spherical Gaussian constraint, named DSG \cite{dsg} is proposed to
offer larger, adaptive step sizes while preserving the original data distribution. 
However, its effectiveness has mainly been shown in image-generation tasks. 
Motivated by DSG, we introduce the Spherical Gaussian constraint to our sampling process, which ensures that the generated hand motion remains within the clean data distribution, resulting in natural and smooth hand motion generation. 

We utilize closed-form solution $\mathcal{H}^{t*}$ that enforces sampling in gradient descent direction while preserving $\mathcal{H}^{t*}$ within noisy data distribution. This can ensure the generated $\mathcal{H}^{0}$ remaining within the clean data manifold:
\begin{equation} 
\mathcal{H}^{t*} = \mu_t - \sqrt{d}\sigma_t \nabla_{\mathcal{H}^t} \mathcal{L}_g(\hat{\mathcal{H}}^0) / ||\nabla_{\mathcal{H}^t} \mathcal{L}_g(\hat{\mathcal{H}}^0)||_2,
\end{equation} 
where $\sigma_t$ represents the variance in time step $t$ and $d$ represents the data dimensions. To enhance generation diversity, we re-weight the direction of $\mathcal{H}^{t*}$ and the sampling point $\mathcal{H}^{t}$ similar to Classifier-free guidance:
\begin{align} \mathbf{D} = \mathcal{H}^{t} + w(\mathcal{H}^{t*}-\mathcal{H}^{t}), \\
\mathcal{H}^{t} = \mu_t + \sqrt{d} \sigma_t \mathbf{D}  / ||\mathbf{D} ||, \end{align} 
where $w$ represents the guidance rate that lies in [0,1] and $\mathcal{D}$ represents the weighted direction.

Since we represent the hand pose using hand joints, a post-optimization process is required to derive the corresponding MANO representation. 
We first calculate the global R  by matching the actual MCP joint positions of the hand with its ideal position in the static state. The MCP joints are crucial for determining the global orientation of the hand because they serve as the primary pivot points for finger movement and are centrally located, providing a stable reference for the overall hand pose. By matching these joints, we can accurately capture the hand's global rotation, ensuring that the subsequent pose adjustments maintain anatomical correctness.
We align the generated joints $\mathbf{J}$ with those from the MANO layer $\mathcal{M}(\cdot)$ via a joint loss $\mathcal{L}_{\text{joint}} = \|\mathbf{J} - \mathcal{M}(\mathbf{H})\|_2$.
To preserve the natural and realistic movement of the hand, we apply pose angle constraints at the end of each iteration. These constraints limit the rotation range of hand joints across different axes, with strict controls specifically on the Y-axis rotation of the MCP joints and the Z-axis rotation of the other joints.

\subsection{Motion Selection via Consistency Score}\label{4.4} 
To ensure generated motions conform to gaze distributions, we evaluate gaze-interaction consistency from local and global perspectives (Fig. \ref{method}b). Locally, we compute a contact map $\mathbf{M}_c = \exp(-\alpha \cdot d(\mathbf{P}_h, \mathbf{P}_o))$ and a gaze map $\mathbf{M}_g = \exp(-\alpha \cdot d(\mathbf{G}, \mathbf{P}_o))$ based on distances from hand vertices $\mathbf{P}_h$ and gaze $\mathbf{G}$ to the object surface $\mathbf{P}_o$. The gaze-contact consistency score is $S_c = \frac{1}{|\mathbf{G}_{\delta}|} \sum_{i \in G_{\delta}} \mathbf{M}_{c_i}$, where $\mathbf{G}_{\delta}$ indexes object points within distance $\delta$ of $\mathbf{G}$. Globally, we quantify trajectory similarity using Dynamic Time Warping (DTW) \cite{seshan2022using}: $\text{DTW}(\mathbf{X}, \mathbf{Y}) = \min_{\pi} \sum_{(i, j) \in \pi} d(x_i, y_j)$, where $\pi$ is the alignment path. The trajectory consistency score is defined as $S_t = \sum_{m \in \{l, r, o\}} \text{DTW}(\mathbf{G},\hat{\mathbf{T}}_m)$. Finally, we select top-$k$ motions $\langle \hat{\mathbf{H}}, \hat{\mathbf{O}} \rangle$ by maximizing the combined objective: $\arg \max_{(\hat{\mathbf{H}}, \hat{\mathbf{O}})} (S_c - \lambda_m S_t)$, where $\lambda_m$ balances the two consistency terms.

\vspace{-2.5ex}
\section{Experiments}
\begin{table*}[!t]
    \centering
    \caption{Quantitative comparison between baselines and our method. The ground truth Diversity is $4.227$ and diversity results are better if the metric is closer to the real distribution.}
    \vspace{-1ex}
    \begin{tabular}{c|ccc|cc|cc|cc}
        \toprule[1.5pt]
        \multirow{2}{*}{\textbf{Methods}} &  \multicolumn{3}{c|}{\textbf{Accuracy}} & \multicolumn{2}{c|}{\textbf{Grasp Quality}} &
        \multicolumn{2}{c|}{\textbf{Generation}} &
        \multicolumn{2}{c}{\textbf{Human Perception}} \\ 
        \cmidrule(r){2-4}
        \cmidrule(r){5-6}
        \cmidrule(r){7-8}
        \cmidrule(r){9-10}
        & MPVPE $\downarrow$ & FOL $\downarrow$ & MPJPE $\downarrow$ & CF$\uparrow$ & PD $\downarrow$  & FID $\downarrow$ & Diversity $\uparrow$  &GMC$\uparrow$&IN$\uparrow$ \\
        \midrule
        
        MDM*& 146.8 & 228.0 & 166.6 & 62.55 & 2.52 & 0.071 & 4.174 &  3.70  &  3.40   \\ 
        
        Text2HOI* & 254.2& 286.7 & 182.2 &63.98 &2.57  & 0.183 &4.134 & 3.05  &  3.75  \\ 
        OMOMO* &- & - & 161.2 & 62.61& 2.49 & 0.051  & 4.121 &  4.15   &   4.05 \\ 
        \midrule
        \textbf{Ours} & \textbf{117.4} & \textbf{146.5} & \textbf{144.8 }& \textbf{68.10}  & \textbf{2.46 }& \textbf{0.044} & \textbf{4.217} & \textbf{4.75}  &\textbf{4.60}     \\ 
        \bottomrule[1.5pt]
    \end{tabular}
    \vspace{-2ex}
    \label{tab:comparison}
\end{table*}

\begin{figure*}[!t]
\centering 
  \begin{minipage}{\textwidth}
    \centering
    \includegraphics[width=1\textwidth]{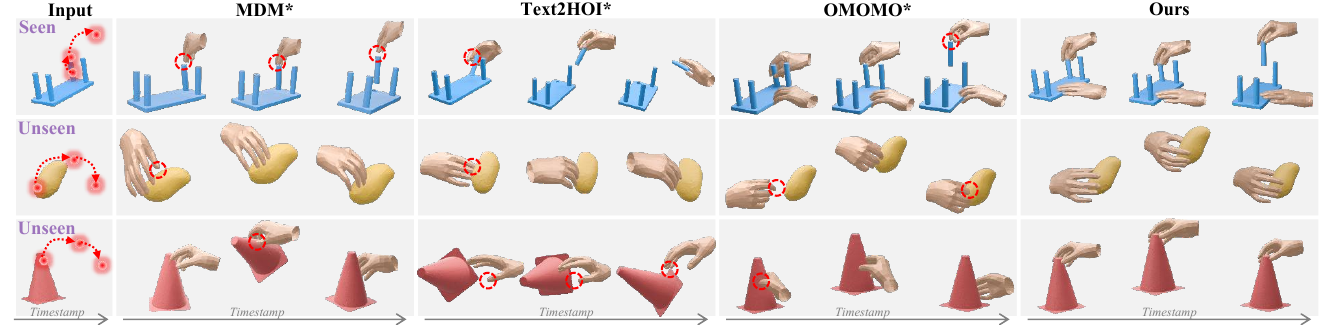}
  \end{minipage}
\caption{Qualitative results between baseline methods and our method.} 
\label{img:comparison}
\vspace{-0.6cm}
\end{figure*}

\subsection{Data Split and Evaluation Metrics}
We divided 1378 sequences into a training set with 1103 sequences and a test set comprising 275 sequences. To mitigate the model's potential overfitting to object-specific traits, the test set exclusively includes objects not represented in the training set. 
Our evaluation metrics includes three parts:
1) motion consistency with gaze and accuracy by evaluating the hand Mean Per Joint Position Error (MPJPE), object Mean Per Vertex Position Error (MPVPE), and the Final Object Location (FOL);  2) the stability and realism of hand-object interactions using the Contact Frame radio (CF) and the Penetration Depth (PD);
3) some basic generation metrics, including FID and Diversity. The FID quantifies the dissimilarity between the real and generated motions, while the diversity measures the variance of different motions. 
Additionally, we conduct a user study in which participants rated the gaze-motion consistency (GMC) and the interactions naturalness (IN) on a five-point scale. Higher scores reflect better motion generation quality.
\vspace{-2ex}
\subsection{Implementation Details}
GHO-Diffusion is built upon a transformer encoder, following the approach of \cite{MDM}, with 1000 denoising steps. We utilize a single NVIDIA GeForce RTX 4090 GPU for training. To simplify the model, we randomly sample 500 points from the surface of the object. The batch size is set to 128, with a learning rate of 1e-5. The latent dimension is uniformly set to 128 for object diffusion and 256 for hand diffusion.
For stage-1, the hyperparameters are set as$\lambda_\alpha = 10$, $\lambda_t = 30$, $\lambda_v = 10$, $\lambda_s = 1$. For stage-2, the hyperparameters are set as $\lambda_\beta = 100$, $\lambda_b = 100$.
For HOI-Manifold guidance, we set $\lambda_k = 100$, $\lambda_c = 1000$,  $\lambda_p = 100$ and guidance rate $w=0.99$. We designed the hyperparameters based on our expertise to balance the weights of each loss component, ensuring they remain within appropriate magnitudes for optimal performance. In motion selection, we set $k=1$ for quantitative experiments and user studies.

\subsection{Baseline Design and Comparison} 

Lacking direct baselines for this novel task, we evaluate our approach against the following adapted methods.
\subsubsection{Text2HOI*} 
The most closely related state-of-the-art work is Text2HOI \cite{text2hoi}, which generates a contact map conditioned on texts and then combines it to produce hand–object motions. Motivated by the relationship between gaze and contact, we adapt this approach by replacing the text encoder with our spatio-temporal gaze encoding to generate the contact map. 
Text2HOI employs a two-stage approach with a contact map as an intermediate representation to refine interaction modeling. However, its focus on single-frame spatial relationships and the use of modified position encoding overlook temporal coherence, which often leads to inconsistent hand–object interactions and noticeable jitter.
\subsubsection{MDM*} 
We adapt MDM \cite{MDM} by replacing its condition encoder with our spatio-temporal gaze encoding and introducing three auxiliary losses: $\mathcal{L}_{\text{obj}} = \sum_{i=1}^L \|\mathbf{T}_o^i -\hat{\mathbf{T}}_o^i\|_2$ and $\mathcal{L}_{\text{hand}} = \sum_{i=1}^L \|\mathbf{T}_h^i -\hat{\mathbf{T}}_h^i\|_2$ for independent translation constraints, and $\mathcal{L}_{\text{offset}} = \sum_{i=1}^L \|(\mathbf{T}_h^i-\mathbf{T}_o^i) -(\hat{\mathbf{T}}_h^i -\hat{\mathbf{T}}_o^i)\|_2$ to regularize the relative hand-object distance. The total training objective is $\mathcal{L}_{\text{MDM*}}= \lambda_\gamma\mathcal{L}_{\text{simple}} + \lambda_o \mathcal{L}_{\text{obj}} + \lambda_h \mathcal{L}_{\text{hand}} + \lambda_f \mathcal{L}_{\text{offset}}$, 
where we set $\{\lambda_\gamma, \lambda_o, \lambda_h, \lambda_f\} = \{20, 20, 20, 30\}$.
MDM employs a simple classifier-free strategy for hand-object motion generation but lacks explicit constraints, resulting in inconsistent and physically implausible hand-object interactions.

\subsubsection{OMOMO*} Object-Conditioned Hand Motion Synthesis is a key component of our method. 
Given the absence of a direct baseline for the full task, we adapt OMOMO \cite{omomo} for comparison by utilizing our Stage-1 object motions as the required conditioning for its hand synthesis process.
OMOMO generates wrist positions and optimizes subsequent trajectories based on the first contact frame to maintain interaction consistency and contact stability. However, this approach overly constrains motion diversity and lacks expressiveness, as wrist position alone is insufficient for fine-grained interaction modeling. In contrast, our method incorporates kinematic, contact, and penetration guidance for more realistic and dynamic interactions. 
As shown in Table \ref{tab:comparison} and Fig. \ref{img:comparison}, our approach significantly outperforms baselines. Note that MPVPE and FOL for OMOMO* are omitted ("-") because it shares our stage-1 object results, making these metrics redundant for comparison.

\begin{table*}[!th]
    \centering
    \scriptsize
    \caption{Quantitative comparison on \textbf{Seen Objects} and \textbf{Unseen Objects}.}
    \vspace{-2ex}

    \begin{tabular}{c|ccc|cc|cc|ccc|cc|cc}
        \toprule[1.5pt]
        \multirow{2}{*}{\textbf{Methods}} &  \multicolumn{7}{c}{\textbf{Seen Objects}} &
        \multicolumn{7}{c}{\textbf{Unseen Objects}}\\
        \cmidrule(r){2-4}
        \cmidrule(r){5-6}
        \cmidrule(r){7-8}
        \cmidrule(r){9-11}
        \cmidrule(r){12-13}
        \cmidrule(r){14-15}
         
        & MPVPE $\downarrow$ & FOL $\downarrow$ & MPJPE $\downarrow$ & CF$\uparrow$ & PD $\downarrow$  &GMC$\uparrow$&IN$\uparrow$ & MPVPE $\downarrow$ & FOL $\downarrow$ & MPJPE $\downarrow$ & CF$\uparrow$ & PD $\downarrow$  &GMC$\uparrow$&IN$\uparrow$ \\
        \midrule
        
        MDM*& 147.5 & 214.4 & 165.8 & 42.30 & 2.17 &  4.05  &  3.55 & 146.6 & 231.8 & 166.8 & 68.19 & 2.62 &  3.40 &  3.35   \\ 
        Text2HOI* &247.5& 275.1 &  179.0&54.66 & 1.98 &  3.25   &  4.25 &256.1& 289.9 &  183.1&66.58 & 2.73 &  3.05   &  3.75 \\

        OMOMO* &- & - & 157.2 & 52.00& 1.95 &  4.30   &   4.2  &- & - & 163.3 & 65.57& 2.65 &  4.00   &   3.60\\ 
        \midrule
        \textbf{Ours} & \textbf{113.1} & \textbf{133.3} & \textbf{146.6 }& \textbf{59.28}  & \textbf{1.92 } & \textbf{4.90}  &\textbf{4.75} & \textbf{118.7} & \textbf{150.2} & \textbf{144.3 }& \textbf{70.55}  & \textbf{2.61 } & \textbf{4.70}  &\textbf{4.55}    \\ 
        \bottomrule[1.5pt]
    \end{tabular}
    \vspace{-4ex}
    \label{tab:seen}
\end{table*}

\begin{table}[!t]
    \centering
    \scriptsize
    \caption{Ablation on Gaze Representation and Encoding.}
    \vspace{-2ex}

    \begin{tabular}{c|cc|c}
    \toprule[1.5pt]
     & MPVPE$\downarrow$ & FOL$\downarrow$& GMC$\uparrow$ \\
    \midrule
    gaze ray & 126.7 &189.2 & 3.95\\
    w/o spatial & 128.4 &193.8& 3.90 \\
    w/o temporal & 119.3 & 163.4&4.15 \\
    w/o post-processing & 134.8 & 179.2 &3.95 \\
    \midrule
   \textbf{ Ours- } & \textbf{117.9} &\textbf{157.6 }&\textbf{4.25} \\

    \bottomrule[1.5pt]
    \end{tabular}
    \vspace{-2ex}
    \label{tab:gaze encoding}

\end{table}

\begin{table}[!t]
    \centering
    \scriptsize
    \caption{Ablation on Different Guidance Strategies.}
    \vspace{-2ex}
    \begin{tabular}{c|c|cc|c}

    \toprule[1.5pt]
      & MPJPE $\downarrow$ & 
  CF$\uparrow$ & PD $\downarrow$ & 
    IN$\uparrow$ \\
    \midrule

    \makecell[c]{w/o guidance} & 155.3 & 64.65 &2.61  &  3.9  \\
    \makecell[c]{na\"ive guidance } & 156.8 & 63.66 & \textbf{2.44} &   3.5  \\
    \midrule
   \textbf{ Ours- }& \textbf{154.3} & \textbf{65.77} & 2.51& \textbf{4.45}   \\
    \bottomrule[1.5pt]
    \end{tabular}
    \vspace{-2ex}
    \label{tab: guidance}
\end{table}

\begin{table}[!t]
    \centering
    \scriptsize
    \caption{Ablation on Motion Selection.}
    \vspace{-2ex}
    \label{tab:gaze_selection}
    \setlength{\tabcolsep}{3.5pt} 
    \begin{tabular}{c|ccc|cc|c|cc}
    \toprule[1.5pt]
    \textbf{} & MPVPE$\downarrow$ & FOL$\downarrow$ &MPJPE$\downarrow$ & CF$\uparrow$&PD$\downarrow$&FID$\downarrow$&
    GMC$\uparrow$&IN$\uparrow$\\
    \midrule
    w/o select & 117.9 & 157.6 & 154.3&65.77&2.51&0.048 &  4.25 &4.45\\
    w/o global & 117.9 & 156.4 & 145.7&67.89&2.47&0.045 & 4.40 &4.55\\
    w/o local &117.5  & 148.9 &149.8 &65.96&2.50&0.046  & 4.35  &4.45 \\
    \midrule
   \textbf{ Ours} & \textbf{117.4} & \textbf{146.5} & \textbf{144.8} &\textbf{68.10}&\textbf{2.46}& \textbf{0.044}&  \textbf{4.75} & \textbf{4.60}\\
    \bottomrule[1.5pt]
    \end{tabular}
    \vspace{-2ex}
\end{table}


\vspace{-2ex}
\subsection{Ablation Study}
\subsubsection{Results on Seen and Unseen Objects}
To assess generalization, our test set (275 sequences) emphasizes novel scenarios: 215 feature entirely unseen objects, while 60 involve seen objects in complex tasks like assembly or chess. These scenarios rigorously test the model's robustness. As shown in Fig. \ref{img:comparison} and Table \ref{tab:seen}, our method consistently produces natural, gaze-aligned motions, outperforming baselines across both seen and unseen categories. This demonstrates superior adaptability to varying interaction complexities without overfitting to object-specific traits.

\subsubsection{Gaze Representation and Encoding}
As the first method to use gaze as a condition for motion generation, we explore the effectiveness of different gaze representations and spatial-temporal feature encoding. The results are in Table \ref{tab:gaze encoding}.
We compare two gaze representations: gaze rays and gaze points. The encoding of gaze points has been detailed in the previous sections. In our framework, the gaze ray is formulated as a unit directional vector. To effectively incorporate this representation, we first process the gaze ray through a linear layer to obtain its embedding. Subsequently, we employ self-attention mechanisms to model its temporal dependencies. Finally, we integrate the gaze ray representation into the diffusion model using the same conditioning strategy as applied to gaze points. However, compared to gaze points, gaze rays provide only directional information and lack precise spatial relationships, weakening spatial modeling and ultimately affecting interaction quality.
The ablation study on spatiotemporal encoding highlights the importance of both temporal and spatial modeling for effective gaze representation. Temporal modeling filters out redundant and noisy data, allowing the model to focus on key gaze information, while spatial modeling captures the relationship between gaze and scene geometry, transforming gaze signals into meaningful interaction cues. The results demonstrate that our method effectively encodes both spatial and temporal features, whereas the absence of either component degrades encoding quality and negatively impacts first-stage performance. Furthermore, we conducted an ablation study on our gaze post-processing pipeline, which confirms its effectiveness in mitigating data noise and improving overall synthesis quality.

\subsubsection{HOI-Manifold Guidance}

Guidance-free methods suffer from contact inconsistencies and penetration errors. While naive guidance strategies can mitigate these issues, they tend to disrupt the learned diffusion process and introduce unnatural motion distortions. In contrast, our HOI-Manifold Guidance optimizes interactions while preserving the learned motion distribution, effectively striking a balance between physical constraints and motion naturalness. As shown in Table \ref{tab: guidance}, this approach yields superior, physically plausible results.

\subsubsection{Motion Selection via Gaze-interaction Consistency Score}

Table \ref{tab:gaze_selection} validates that our consistency score selects motions better aligned with gaze cues. Specifically, local selection refines fine-grained hand-object contact details, significantly improving MPJPE, CF, and PD. Meanwhile, global selection enhances overall trajectory consistency with spatial gaze cues, leading to notable gains in MPVPE and FOL.
\begin{table}[!t]
    \centering
    \scriptsize
    \caption{Performance of Reconstruction Baselines.}
    \vspace{-2ex}
    \label{tab:recon}
    \setlength{\tabcolsep}{3pt}
    \begin{tabular}{c|c|cc|c}
    \toprule[1.5pt]
        Splits & Methods & MRRPE$\downarrow$& MPJPE$\downarrow$& SR$\uparrow$\\
         \midrule
         \multirow{3}{*}{Val.}  & Arctic-SF \cite{arctic} &  111.2/43.5&\textbf{23.7}&20.3\\
         & Arctic-LSTM \cite{arctic}  &114.9/45.2&24.0&22.5\\
         & Kypt Trans. \cite{kypt} & \textbf{108.9/42.6}&32.6&\textbf{28.4}\\
          \midrule
         \multirow{3}{*}{Test}  & Arctic-SF \cite{arctic}  & \textbf{65.1}/42.6&\textbf{22.8}&29.7\\
         & Arctic-LSTM \cite{arctic} & 68.9/\textbf{42.4}&23.1&33.0\\
         & Kypt Trans. \cite{kypt} & 111.4/42.5 &27.6&\textbf{44.9}\\
    \bottomrule[1.5pt]
    \end{tabular}
    \vspace{-5ex}
\end{table}
\section{Task: Hand-Object Motion Reconstruction}\label{rec}
Leveraging rigorous data collection and precise cross-coordinate alignment, GazeHOI enables high-quality 3D hand-object reconstruction from visual cues. Given an interaction video and object mesh, the task aims to reconstruct continuous and natural hand-object motions for each frame.

\textbf{Data and Metrics:} We utilize 12 side views (3.0M images), split by subject: 8 for training, 1 for validation, and 1 for testing. Evaluation metrics include Mean Relative-Root Position Error (MRRPE) for relative root positions, local Mean Per Joint Position Error (MPJPE), and Success Rate \cite{arctic} (percentage of predicted object vertices with L2 error $< 5\%$ of the object's diameter).

\textbf{Baselines and Results:} To establish a benchmark, we evaluate two baselines: (1) ArcticNet \cite{arctic}, including an image-based version (SF) and a video-based version (LSTM) for temporal reasoning; and (2) Keypoint Transformer \cite{kypt}, an image-based model effective for joint identification across challenging datasets like InterHand2.6M \cite{interhand} and HO-3D \cite{ho3d}. Table \ref{tab:recon} presents the quantitative results, validating GazeHOI's utility for reconstruction tasks beyond synthesis.

\vspace{-2ex}
\section{Conclusion}

In this paper, we introduced Gaze-guided Hand-Object Interaction  Synthesis and GazeHOI, the first dataset capturing simultaneous 3D gaze, hand, and object interactions. We proposed GHO-Diffusion, a stacked diffusion model with HOI-Manifold Guidance to ensure physically plausible sampling. Furthermore, we leveraged gaze-contact and trajectory consistency to refine the motions. Experimental results validate the effectiveness of our method and the unique contributions of our dataset.
Despite its diversity, GazeHOI currently excludes articulated objects, which present unique interaction challenges. Future work will extend this research to incorporate articulated interactions and multi-modal inputs, such as text and electromyography (EMG) data, to generate more context-aware interactive motions.

\vspace{-2ex}
\bibliographystyle{IEEEtran}
\bibliography{tmm}

@InProceedings{ho3d,
author = {Hampali, Shreyas and Rad, Mahdi and Oberweger, Markus and Lepetit, Vincent},
title = {HOnnotate: A Method for 3D Annotation of Hand and Object Poses},
booktitle = {Proceedings of the IEEE/CVF Conference on Computer Vision and Pattern Recognition (CVPR)},
month = {June},
year = {2020}
}

@InProceedings{dexycb,
    author    = {Chao, Yu-Wei and Yang, Wei and Xiang, Yu and Molchanov, Pavlo and Handa, Ankur and Tremblay, Jonathan and Narang, Yashraj S. and Van Wyk, Karl and Iqbal, Umar and Birchfield, Stan and Kautz, Jan and Fox, Dieter},
    title     = {DexYCB: A Benchmark for Capturing Hand Grasping of Objects},
    booktitle = {Proceedings of the IEEE/CVF Conference on Computer Vision and Pattern Recognition (CVPR)},
    month     = {June},
    year      = {2021},
    pages     = {9044-9053}
}

@InProceedings{h2o,
    author    = {Kwon, Taein and Tekin, Bugra and St\"uhmer, Jan and Bogo, Federica and Pollefeys, Marc},
    title     = {H2O: Two Hands Manipulating Objects for First Person Interaction Recognition},
    booktitle = {Proceedings of the IEEE/CVF International Conference on Computer Vision (ICCV)},
    month     = {October},
    year      = {2021},
    pages     = {10138-10148}
}

@InProceedings{oakink,
    author    = {Yang, Lixin and Li, Kailin and Zhan, Xinyu and Wu, Fei and Xu, Anran and Liu, Liu and Lu, Cewu},
    title     = {OakInk: A Large-Scale Knowledge Repository for Understanding Hand-Object Interaction},
    booktitle = {Proceedings of the IEEE/CVF Conference on Computer Vision and Pattern Recognition (CVPR)},
    month     = {June},
    year      = {2022},
    pages     = {20953-20962}
}

@InProceedings{hoi4d,
    author    = {Liu, Yunze and Liu, Yun and Jiang, Che and Lyu, Kangbo and Wan, Weikang and Shen, Hao and Liang, Boqiang and Fu, Zhoujie and Wang, He and Yi, Li},
    title     = {HOI4D: A 4D Egocentric Dataset for Category-Level Human-Object Interaction},
    booktitle = {Proceedings of the IEEE/CVF Conference on Computer Vision and Pattern Recognition (CVPR)},
    month     = {June},
    year      = {2022},
    pages     = {21013-21022}
}

@InProceedings{arctic,
    author    = {Fan, Zicong and Taheri, Omid and Tzionas, Dimitrios and Kocabas, Muhammed and Kaufmann, Manuel and Black, Michael J. and Hilliges, Otmar},
    title     = {ARCTIC: A Dataset for Dexterous Bimanual Hand-Object Manipulation},
    booktitle = {Proceedings of the IEEE/CVF Conference on Computer Vision and Pattern Recognition (CVPR)},
    month     = {June},
    year      = {2023},
    pages     = {12943-12954}
}

@article{taco, title={TACO: Benchmarking Generalizable Bimanual Tool-ACtion-Object Understanding}, author={Liu, Yun and Yang, Haolin and Si, Xu and Liu, Ling and Li, Zipeng and Zhang, Yuxiang and Liu, Yebin and Yi, Li}, journal={arXiv preprint arXiv:2401.08399}, year={2024} }

@article{Lou2024PredictingRG,
  title={Predicting Radiologists' Gaze With Computational Saliency Models in Mammogram Reading},
  author={Jianxun Lou and Hanhe Lin and Philippa Young and Richard White and Zelei Yang and Susan Cheng Shelmerdine and David Marshall and Emiliano Spezi and Marco Palombo and Hantao Liu},
  journal={IEEE Transactions on Multimedia},
  year={2024},
  volume={26},
  pages={256-269}
}

@article{Zhang2024DomainConsistentAU,
  title={Domain-Consistent and Uncertainty-Aware Network for Generalizable Gaze Estimation},
  author={Sihui Zhang and Yi Tian and Yilei Zhang and Mei Tian and Yaping Huang},
  journal={IEEE Transactions on Multimedia},
  year={2024},
  volume={26},
  pages={6996-7011}}

@article{Lv2021ImprovingDG,
  title={Improving Driver Gaze Prediction With Reinforced Attention},
  author={Kai Lv and Hao Sheng and Zhang Xiong and Wei Li and Liang Zheng},
  journal={IEEE Transactions on Multimedia},
  year={2021},
  volume={23},
  pages={4198-4207}
}

@inproceedings{ge1, title={L2cs-net: Fine-grained gaze estimation in unconstrained environments}, author={Abdelrahman, Ahmed A and Hempel, Thorsten and Khalifa, Aly and Al-Hamadi, Ayoub and Dinges, Laslo}, booktitle={2023 8th International Conference on Frontiers of Signal Processing (ICFSP)}, pages={98--102}, year={2023}, organization={IEEE} }

@article{ge2, title={End-to-End Video Gaze Estimation via Capturing Head-Face-Eye Spatial-Temporal Interaction Context}, author={Guan, Yiran and Chen, Zhuoguang and Zeng, Wenzheng and Cao, Zhiguo and Xiao, Yang}, journal={IEEE Signal Processing Letters}, volume={30}, pages={1687--1691}, year={2023}, publisher={IEEE} }

@inproceedings{ge3, title={Weakly-supervised physically unconstrained gaze estimation}, author={Kothari, Rakshit and De Mello, Shalini and Iqbal, Umar and Byeon, Wonmin and Park, Seonwook and Kautz, Jan}, booktitle={Proceedings of the IEEE/CVF Conference on Computer Vision and Pattern Recognition}, pages={9980--9989}, year={2021} }

@inproceedings{ge4, title={Rt-gene: Real-time eye gaze estimation in natural environments}, author={Fischer, Tobias and Chang, Hyung Jin and Demiris, Yiannis}, booktitle={Proceedings of the European conference on computer vision (ECCV)}, pages={334--352}, year={2018} }

@article{Chen20233DFR,
  title={3D Face Reconstruction and Gaze Tracking in the HMD for Virtual Interaction},
  author={Shu-Yu Chen and Yu-Kun Lai and Shi-hong Xia and Paul L. Rosin and Lin Gao},
  journal={IEEE Transactions on Multimedia},
  year={2023},
  volume={25},
  pages={3166-3179}}

@article{mogaze,
  title={Mogaze: A dataset of full-body motions that includes workspace geometry and eye-gaze},
  author={Kratzer, Philipp and Bihlmaier, Simon and Midlagajni, Niteesh Balachandra and Prakash, Rohit and Toussaint, Marc and Mainprice, Jim},
  journal={IEEE Robotics and Automation Letters},
  volume={6},
  number={2},
  pages={367--373},
  year={2020},
  publisher={IEEE}
}

@inproceedings{meccano,
  title={The meccano dataset: Understanding human-object interactions from egocentric videos in an industrial-like domain},
  author={Ragusa, Francesco and Furnari, Antonino and Livatino, Salvatore and Farinella, Giovanni Maria},
  booktitle={Proceedings of the IEEE/CVF Winter Conference on Applications of Computer Vision},
  pages={1569--1578},
  year={2021}
}

@article{actionsense,
  title={ActionSense: A multimodal dataset and recording framework for human activities using wearable sensors in a kitchen environment},
  author={DelPreto, Joseph and Liu, Chao and Luo, Yiyue and Foshey, Michael and Li, Yunzhu and Torralba, Antonio and Matusik, Wojciech and Rus, Daniela},
  journal={Advances in Neural Information Processing Systems},
  volume={35},
  pages={13800--13813},
  year={2022}
}

@inproceedings{gimo,
  title={Gimo: Gaze-informed human motion prediction in context},
  author={Zheng, Yang and Yang, Yanchao and Mo, Kaichun and Li, Jiaman and Yu, Tao and Liu, Yebin and Liu, C Karen and Guibas, Leonidas J},
  booktitle={European Conference on Computer Vision},
  pages={676--694},
  year={2022},
  organization={Springer}
}

@inproceedings{egobody,
  title={Egobody: Human body shape and motion of interacting people from head-mounted devices},
  author={Zhang, Siwei and Ma, Qianli and Zhang, Yan and Qian, Zhiyin and Kwon, Taein and Pollefeys, Marc and Bogo, Federica and Tang, Siyu},
  booktitle={European Conference on Computer Vision},
  pages={180--200},
  year={2022},
  organization={Springer}
}

@inproceedings{holoassist,
  title={Holoassist: an egocentric human interaction dataset for interactive ai assistants in the real world},
  author={Wang, Xin and Kwon, Taein and Rad, Mahdi and Pan, Bowen and Chakraborty, Ishani and Andrist, Sean and Bohus, Dan and Feniello, Ashley and Tekin, Bugra and Frujeri, Felipe Vieira and others},
  booktitle={Proceedings of the IEEE/CVF International Conference on Computer Vision},
  pages={20270--20281},
  year={2023}
}

@inproceedings{aria,
  title={Aria digital twin: A new benchmark dataset for egocentric 3d machine perception},
  author={Pan, Xiaqing and Charron, Nicholas and Yang, Yongqian and Peters, Scott and Whelan, Thomas and Kong, Chen and Parkhi, Omkar and Newcombe, Richard and Ren, Yuheng Carl},
  booktitle={Proceedings of the IEEE/CVF International Conference on Computer Vision},
  pages={20133--20143},
  year={2023}
}

@article{gazedrive, title={G-MEMP: Gaze-Enhanced Multimodal Ego-Motion Prediction in Driving}, author={Eren Akbiyik, M and Savov, Nedko and Pani Paudel, Danda and Popovic, Nikola and Vater, Christian and Hilliges, Otmar and Van Gool, Luc and Wang, Xi}, journal={arXiv e-prints}, pages={arXiv--2312}, year={2023} }

@article{gazemotion,
  title={GazeMotion: Gaze-guided Human Motion Forecasting},
  author={Hu, Zhiming and Schmitt, Syn and Haeufle, Daniel and Bulling, Andreas},
  journal={arXiv preprint arXiv:2403.09885},
  year={2024}}

@article{gazemodiff,
  title={GazeMoDiff: Gaze-guided Diffusion Model for Stochastic Human Motion Prediction},
  author={Yan, Haodong and Hu, Zhiming and Schmitt, Syn and Bulling, Andreas},
  journal={arXiv preprint arXiv:2312.12090},
  year={2023}
}

@inproceedings{liu2020forecasting, title={Forecasting human-object interaction: joint prediction of motor attention and actions in first person video}, author={Liu, Miao and Tang, Siyu and Li, Yin and Rehg, James M}, booktitle={Computer Vision--ECCV 2020: 16th European Conference, Glasgow, UK, August 23--28, 2020, Proceedings, Part I 16}, pages={704--721}, year={2020}, organization={Springer} }

@article{MDM,
  title={Human motion diffusion model},
  author={Tevet, Guy and Raab, Sigal and Gordon, Brian and Shafir, Yonatan and Cohen-Or, Daniel and Bermano, Amit H},
  journal={arXiv preprint arXiv:2209.14916},
  year={2022}
}

@article{mano,
      title = {Embodied Hands: Modeling and Capturing Hands and Bodies Together},
      author = {Romero, Javier and Tzionas, Dimitrios and Black, Michael J.},
      journal = {ACM Transactions on Graphics, (Proc. SIGGRAPH Asia)},
      volume = {36},
      number = {6},
      series = {245:1--245:17},
      month = nov,
      year = {2017},
      month_numeric = {11}
}

@article{mediapipe, author = {Camillo Lugaresi and Jiuqiang Tang and Hadon Nash and Chris McClanahan and Esha Uboweja and Michael Hays and Fan Zhang and Chuo{-}Ling Chang and Ming Guang Yong and Juhyun Lee and Wan{-}Teh Chang and Wei Hua and Manfred Georg and Matthias Grundmann}, title = {MediaPipe: {A} Framework for Building Perception Pipelines}, journal = {CoRR}, volume = {abs/1906.08172}, year = {2019}, url = {http://arxiv.org/abs/1906.08172}, eprinttype = {arXiv}, eprint = {1906.08172}, bibsource = {dblp computer science bibliography, https://dblp.org} }

@article{con1, author = {Werner X. Schneider}, title = {VAM: A neuro-cognitive model for visual attention control of segmentation, object recognition, and space-based motor action}, journal = {Visual Cognition}, volume = {2}, number = {2-3}, pages = {331-376}, year = {1995}, publisher = {Routledge}, doi = {10.1080/13506289508401737}, URL = { https://doi.org/10.1080/13506289508401737 } }

@article{con3, title={Information-seeking, curiosity, and attention: computational and neural mechanisms}, author={Gottlieb, Jacqueline and Oudeyer, Pierre-Yves and Lopes, Manuel and Baranes, Adrien}, journal={Trends in cognitive sciences}, volume={17}, number={11}, pages={585--593}, year={2013}, publisher={Elsevier} }

@article{con4grasp, title={The influence of object shape and center of mass on grasp and gaze}, author={Desanghere, Loni and Marotta, Jonathan J}, journal={Frontiers in psychology}, volume={6}, pages={1537}, year={2015}, publisher={Frontiers Media SA} }

@article{qi2017pointnet++, title={Pointnet++: Deep hierarchical feature learning on point sets in a metric space}, author={Qi, Charles Ruizhongtai and Yi, Li and Su, Hao and Guibas, Leonidas J}, journal={Advances in neural information processing systems}, volume={30}, year={2017} }

@inproceedings{liu2023contactgen, title={ContactGen: Generative Contact Modeling for Grasp Generation}, author={Liu, Shaowei and Zhou, Yang and Yang, Jimei and Gupta, Saurabh and Wang, Shenlong}, booktitle={Proceedings of the IEEE/CVF International Conference on Computer Vision}, pages={20609--20620}, year={2023} }

@inproceedings{ho_vae_1,
  title={GOAL: Generating 4D whole-body motion for hand-object grasping},
  author={Taheri, Omid and Choutas, Vasileios and Black, Michael J and Tzionas, Dimitrios},
  booktitle={Proceedings of the IEEE/CVF Conference on Computer Vision and Pattern Recognition},
  pages={13263--13273},
  year={2022}
}

@misc{ho_vae_2,
      title={IMos: Intent-Driven Full-Body Motion Synthesis for Human-Object Interactions}, 
      author={Anindita Ghosh and Rishabh Dabral and Vladislav Golyanik and Christian Theobalt and Philipp Slusallek},
      year={2023},
      eprint={2212.07555},
      archivePrefix={arXiv},
      primaryClass={cs.CV}
}

@inproceedings{ho_diff_1,
  title={Guided Motion Diffusion for Controllable Human Motion Synthesis},
  author={Karunratanakul, Korrawe and Preechakul, Konpat and Suwajanakorn, Supasorn and Tang, Siyu},
  booktitle={Proceedings of the IEEE/CVF International Conference on Computer Vision},
  pages={2151--2162},
  year={2023}
}

@article{hdo_1,
  title={Contact2Grasp: 3D Grasp Synthesis via Hand-Object Contact Constraint},
  author={Li, Haoming and Lin, Xinzhuo and Zhou, Yang and Li, Xiang and Huo, Yuchi and Chen, Jiming and Ye, Qi},
  journal={arXiv preprint arXiv:2210.09245},
  year={2022}
}

@inproceedings{hdo_2,
  title={Hand-object contact consistency reasoning for human grasps generation},
  author={Jiang, Hanwen and Liu, Shaowei and Wang, Jiashun and Wang, Xiaolong},
  booktitle={Proceedings of the IEEE/CVF International Conference on Computer Vision},
  pages={11107--11116},
  year={2021}
}

@article{hdo_3, 
  title={Manipnet: neural manipulation synthesis with a hand-object spatial representation},
  author={Zhang, He and Ye, Yuting and Shiratori, Takaaki and Komura, Taku},
  journal={ACM Transactions on Graphics (ToG)},
  volume={40},
  number={4},
  pages={1--14},
  year={2021},
  publisher={ACM New York, NY, USA}
}

@inproceedings{grip,
  title={GRIP: Generating interaction poses using spatial cues and latent consistency},
  author={Taheri, Omid and Zhou, Yi and Tzionas, Dimitrios and Zhou, Yang and Ceylan, Duygu and Pirk, Soren and Black, Michael J},
  booktitle={2024 International Conference on 3D Vision (3DV)},
  pages={933--943},
  year={2024},
  organization={IEEE}
}

@inproceedings{gears,
  title={GEARS: Local Geometry-aware Hand-object Interaction Synthesis},
  author={Zhou, Keyang and Bhatnagar, Bharat Lal and Lenssen, Jan Eric and Pons-Moll, Gerard},
  booktitle={Proceedings of the IEEE/CVF Conference on Computer Vision and Pattern Recognition},
  pages={20634--20643},
  year={2024}
}

@inproceedings{hdo_4,
  title={CAMS: CAnonicalized Manipulation Spaces for Category-Level Functional Hand-Object Manipulation Synthesis},
  author={Zheng, Juntian and Zheng, Qingyuan and Fang, Lixing and Liu, Yun and Yi, Li},
  booktitle={Proceedings of the IEEE/CVF Conference on Computer Vision and Pattern Recognition},
  pages={585--594},
  year={2023}
}

@inproceedings{hdo_6,
  title={D-grasp: Physically plausible dynamic grasp synthesis for hand-object interactions},
  author={Christen, Sammy and Kocabas, Muhammed and Aksan, Emre and Hwangbo, Jemin and Song, Jie and Hilliges, Otmar},
  booktitle={Proceedings of the IEEE/CVF Conference on Computer Vision and Pattern Recognition},
  pages={20577--20586},
  year={2022}
}

@article{dsg,
  title={Guidance with Spherical Gaussian Constraint for Conditional Diffusion},
  author={Yang, Lingxiao and Ding, Shutong and Cai, Yifan and Yu, Jingyi and Wang, Jingya and Shi, Ye},
  journal={arXiv preprint arXiv:2402.03201},
  year={2024}
}

@article{omomo,
  title={Object motion guided human motion synthesis},
  author={Li, Jiaman and Wu, Jiajun and Liu, C Karen},
  journal={ACM Transactions on Graphics (TOG)},
  volume={42},
  number={6},
  pages={1--11},
  year={2023},
  publisher={ACM New York, NY, USA}
}

@inproceedings{contactopt,
  title={Contactopt: Optimizing contact to improve grasps},
  author={Grady, Patrick and Tang, Chengcheng and Twigg, Christopher D and Vo, Minh and Brahmbhatt, Samarth and Kemp, Charles C},
  booktitle={Proceedings of the IEEE/CVF Conference on Computer Vision and Pattern Recognition},
  pages={1471--1481},
  year={2021}
}

@inproceedings{text2hoi,
  title={Text2HOI: Text-guided 3D Motion Generation for Hand-Object Interaction},
  author={Cha, Junuk and Kim, Jihyeon and Yoon, Jae Shin and Baek, Seungryul},
  booktitle={Proceedings of the IEEE/CVF Conference on Computer Vision and Pattern Recognition},
  pages={1577--1585},
  year={2024}
}

@article{diffh2o,
  title={Diffh2o: Diffusion-based synthesis of hand-object interactions from textual descriptions},
  author={Christen, Sammy and Hampali, Shreyas and Sener, Fadime and Remelli, Edoardo and Hodan, Tomas and Sauser, Eric and Ma, Shugao and Tekin, Bugra},
  journal={arXiv:2403.17827},
  year={2024}
}

@article{zhao2023m,
  title={I'M HOI: Inertia-aware Monocular Capture of 3D Human-Object Interactions},
  author={Zhao, Chengfeng and Zhang, Juze and Du, Jiashen and Shan, Ziwei and Wang, Junye and Yu, Jingyi and Wang, Jingya and Xu, Lan},
  journal={arXiv preprint arXiv:2312.08869},
  year={2023}
}

@InProceedings{neuraldome,
    author    = {Zhang, Juze and Luo, Haimin and Yang, Hongdi and Xu, Xinru and Wu, Qianyang and Shi, Ye and Yu, Jingyi and Xu, Lan and Wang, Jingya},
    title     = {NeuralDome: A Neural Modeling Pipeline on Multi-View Human-Object Interactions},
    booktitle = {Proceedings of the IEEE/CVF Conference on Computer Vision and Pattern Recognition (CVPR)},
    month     = {June},
    year      = {2023},
    pages     = {8834-8845}
}

@INPROCEEDINGS{kypt,
  author={Hampali, Shreyas and Sarkar, Sayan Deb and Rad, Mahdi and Lepetit, Vincent},
  booktitle={2022 IEEE/CVF Conference on Computer Vision and Pattern Recognition (CVPR)}, 
  title={Keypoint Transformer: Solving Joint Identification in Challenging Hands and Object Interactions for Accurate 3D Pose Estimation}, 
  year={2022},
  volume={},
  number={},
  pages={11080-11090},
  doi={10.1109/CVPR52688.2022.01081}}

@inproceedings{zhan2024oakink2,
  title={OAKINK2: A Dataset of Bimanual Hands-Object Manipulation in Complex Task Completion},
  author={Zhan, Xinyu and Yang, Lixin and Zhao, Yifei and Mao, Kangrui and Xu, Hanlin and Lin, Zenan and Li, Kailin and Lu, Cewu},
  booktitle={Proceedings of the IEEE/CVF Conference on Computer Vision and Pattern Recognition},
  pages={445--456},
  year={2024}
}

@article{seshan2022using,
  title={Using Machine Learning to Augment Dynamic Time Warping Based Signal Classification},
  author={Seshan, Arvind},
  journal={arXiv preprint arXiv:2206.07200},
  year={2022}
}

@inproceedings{interhand,
  title={Interhand2. 6m: A dataset and baseline for 3d interacting hand pose estimation from a single rgb image},
  author={Moon, Gyeongsik and Yu, Shoou-I and Wen, He and Shiratori, Takaaki and Lee, Kyoung Mu},
  booktitle={Computer Vision--ECCV 2020: 16th European Conference, Glasgow, UK, August 23--28, 2020, Proceedings, Part XX 16},
  pages={548--564},
  year={2020},
  organization={Springer}
}

\end{document}